\documentclass{article}

\usepackage[preprint]{neurips_2025}

\usepackage[utf8]{inputenc} % allow utf-8 input
\usepackage[T1]{fontenc}    % use 8-bit T1 fonts
\usepackage{hyperref}       % hyperlinks
\usepackage{url}            % simple URL typesetting
\usepackage{booktabs}       % professional-quality tables
\usepackage{amsfonts}       % blackboard math symbols
\usepackage{nicefrac}       % compact symbols for 1/2, etc.
\usepackage{microtype}      % microtypography
\usepackage{xcolor}         % colors
\usepackage{graphicx}
\usepackage{enumitem}
\usepackage{multirow}
\usepackage{tabularx}
\usepackage{float}
\usepackage{caption}
\usepackage{booktabs}
\usepackage{wrapfig}
\usepackage[font=footnotesize]{caption}
\title{Towards Fine-Grained Recognition with Large Visual Language Models: Benchmark and Optimization Strategies}

\author{
\begin{minipage}{0.5\linewidth}
\centering
{\bfseries Cong Pang}\thanks{Equal contribution.}\\
{\normalfont ShanghaiTech University}\\
{\normalfont \texttt{pangcong2022@shanghaitech.edu.cn}}
\end{minipage}
\hfill
\begin{minipage}{0.5\linewidth}
\centering
{\bfseries Hongtao Yu}\footnotemark[1]\\
{\normalfont Southeast University}\\
{\normalfont \texttt{yuht@njust.edu.cn}}
\end{minipage}
\\[1.8em]
\begin{minipage}{0.5\linewidth}
\centering
{\bfseries Zixuan Chen}\\
{\normalfont SenseTime Research}\\
{\normalfont \texttt{chenzixuan3@sensetime.com}}
\end{minipage}
\hfill
\begin{minipage}{0.5\linewidth}
\centering
{\bfseries Lewei Lu}\\
{\normalfont SenseTime Research}\\
{\normalfont \texttt{luotto@sensetime.com}}
\end{minipage}
\\[1.8em]
\begin{minipage}{0.99\linewidth}
\centering
{\bfseries Xin Lou}\thanks{Corresponding author.}\\
{\normalfont ShanghaiTech University}\\
{\normalfont \texttt{louxin@shanghaitech.edu.cn}}
\end{minipage}
}

\begin{document}

\maketitle

\begin{abstract}
Large Vision Language Models (LVLMs) have made remarkable progress, enabling sophisticated vision-language interaction and dialogue applications. However, existing benchmarks primarily focus on reasoning tasks, often neglecting fine-grained recognition, which is crucial for practical application scenarios. To address this gap, we introduce the Fine-grained Recognition Open World (FROW) benchmark, designed for detailed evaluation of LVLMs with GPT-4o. On the basis of that, we propose a novel optimization strategy from two perspectives: \textit{data construction} and \textit{training process}, to improve the performance of LVLMs. Our dataset includes mosaic data, which combines multiple short-answer responses, and open-world data, generated from real-world questions and answers using GPT-4o, creating a comprehensive framework for evaluating fine-grained recognition in LVLMs. Experiments show that mosaic data improves category recognition accuracy by 1\% and open-world data boosts FROW benchmark accuracy by 10\%-20\% and content accuracy by 6\%-12\%. Meanwhile, incorporating fine-grained data into the pre-training phase can improve the model's category recognition accuracy by up to 10\%. The benchmark will be available at \url{https://github.com/pc-inno/FROW}.
\end{abstract}

\section{Introduction}
\begin{figure}[t]
    \centering
    \includegraphics[width=0.75\linewidth]{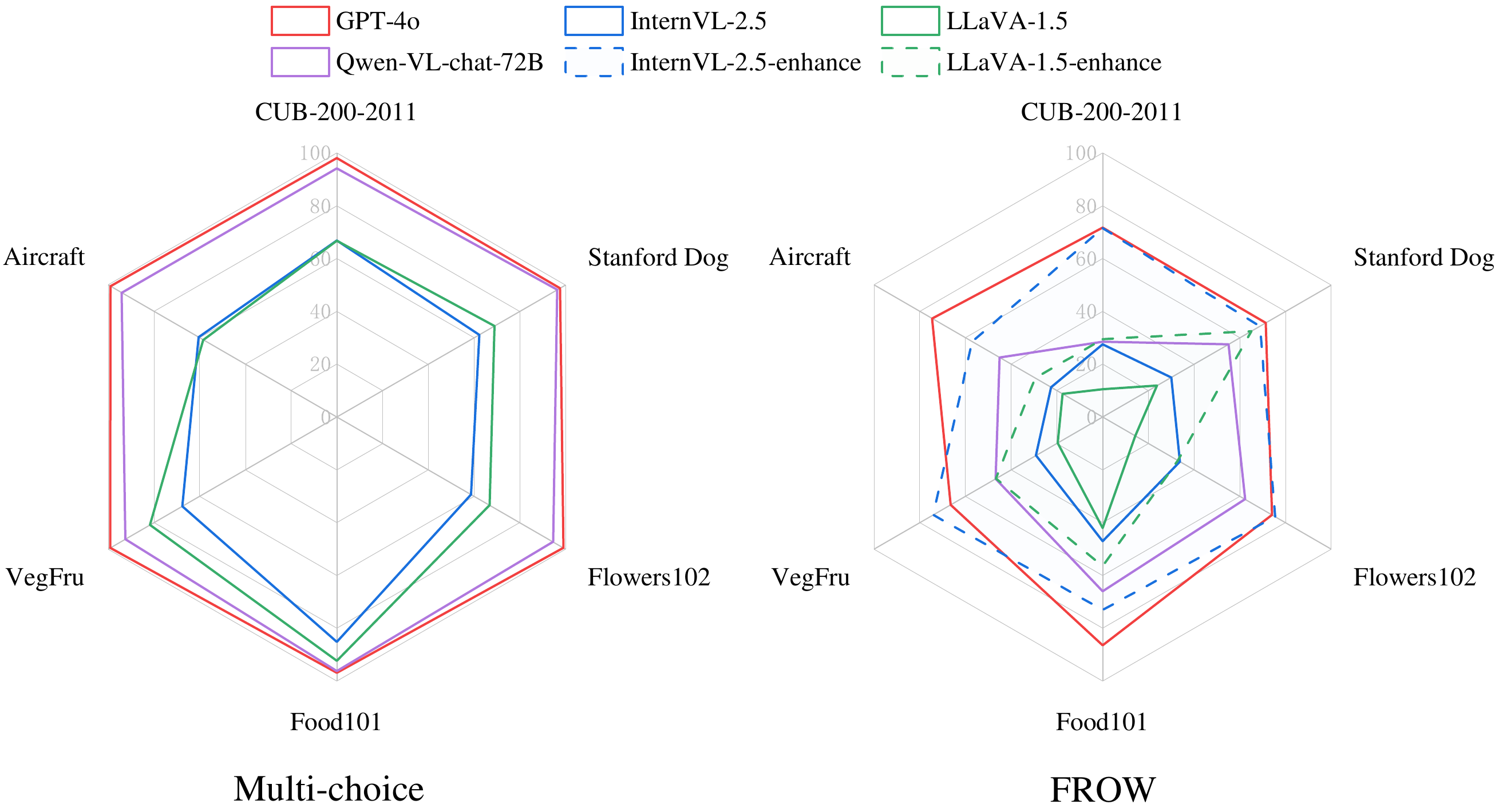}
    % \vspace{-2.1em}
    \caption{The left panel depicts the model's recognition accuracy on fine-grained multiple-choice questions, while the right panel showcases its accuracy on the FROW benchmark. Detailed scores are provided in Table \ref{tab:scores}. Dashed lines represent the recognition accuracy of LVLMs optimized using the proposed strategy, highlighting performance improvements for both InternVL and LLaVA. The images used in the evaluation are drawn from six fine-grained datasets: FGVC-Aircraft \cite{aircraft}, Caltech-UCSD Birds-200-2011 \cite{CUB}, Food-101 \cite{food101}, Stanford Dogs \cite{stanforddog}, Oxford Flowers-102 \cite{nilsback2008automated}, and VegFru \cite{hou2017vegfru}.}
    \vspace{-1.2em}  % 这里调整caption和正文之间的距离
    \label{fig:scores}
\end{figure}

Large Language Models (LLMs) have demonstrated exceptional performance in open-domain language tasks, marking significant strides toward the realization of Artificial General Intelligence (AGI). Similarly, advancements in Large Vision-Language Models (LVLMs) \cite{Qwen-VL, internvl, anthropic2024claude, hurst2024gpt, liu2023llava, liu2024llavanext, team2024gemini} have enabled sophisticated vision-language interactions and complex dialogue capabilities.
To evaluate these models, a variety of benchmarks have been introduced, spanning from general-purpose to domain-specific tasks~\cite{DocVQA, ChartQA, mathew2022infographicvqa, wu2024qbench, MM1, zhao2024videoniah, lu2024mathvista}. However, few evaluations focus on fine-grained image tasks—a critical aspect of computer vision—which involve distinguishing objects among multiple subordinate categories. Current benchmarks~\cite{VLMClassifier, geigle-etal-2024-african} predominantly evaluate fine-grained capabilities using multiple-choice questions, simplifying the task by limiting the search space. Consequently, the true extent of LVLMs' fine-grained recognition abilities remains unclear. For instance, as illustrated in Figure~\ref{fig:scores}, GPT-4o achieved near-perfect accuracy in multiple-choice fine-grained recognition tasks, highlighting the limitations of existing evaluations.

To address the gap, we develop a more challenging \textbf{F}ine-grained \textbf{R}ecognition \textbf{O}pen-\textbf{W}orld benchmark (\textbf{FROW}) based on images and Wikipedia content. To thoroughly assess LVLMs and their fine-grained recognition capabilities, we employ expert models to evaluate various models on this benchmark. Composed entirely of open-ended questions, FROW requires models to identify objects in images accurately before providing correct answers. The benchmark assesses responses using two primary metrics: \textit{recognition accuracy} and \textit{content accuracy}. Recognition accuracy evaluates whether a model correctly identifies objects in images, whereas content accuracy gauges whether the model’s responses are factually correct.
As shown in Figure~\ref{fig:scores}, the results demonstrate that both proprietary and open-source LVLMs exhibit significant deficiencies in fine-grained recognition and domain-specific knowledge. These findings highlight the critical need for improving LVLMs' performance in such tasks to enable meaningful reasoning. Misidentifying objects renders subsequent reasoning invalid, emphasizing the foundational importance of fine-grained recognition, as also highlighted by~\cite{VLMClassifier}.
% 有其他一些工作也强调了这个重要性
Given these limitations, we further propose dedicated optimization strategies. Previous research~\cite{VLMClassifier} attributes the inadequate performance of LVLMs in fine-grained tasks primarily to the lack of sufficient fine-grained data. Our study reaffirms the pivotal role of data in advancing these capabilities. Moreover, limited attention has been devoted to fine-grained recognition during training, leaving a significant gap in model development. To address this, we present a comprehensive approach to optimize LVLM training and enhance their fine-grained recognition performance.

\begin{figure*}[t]
    \centering
    \includegraphics[width=\linewidth]{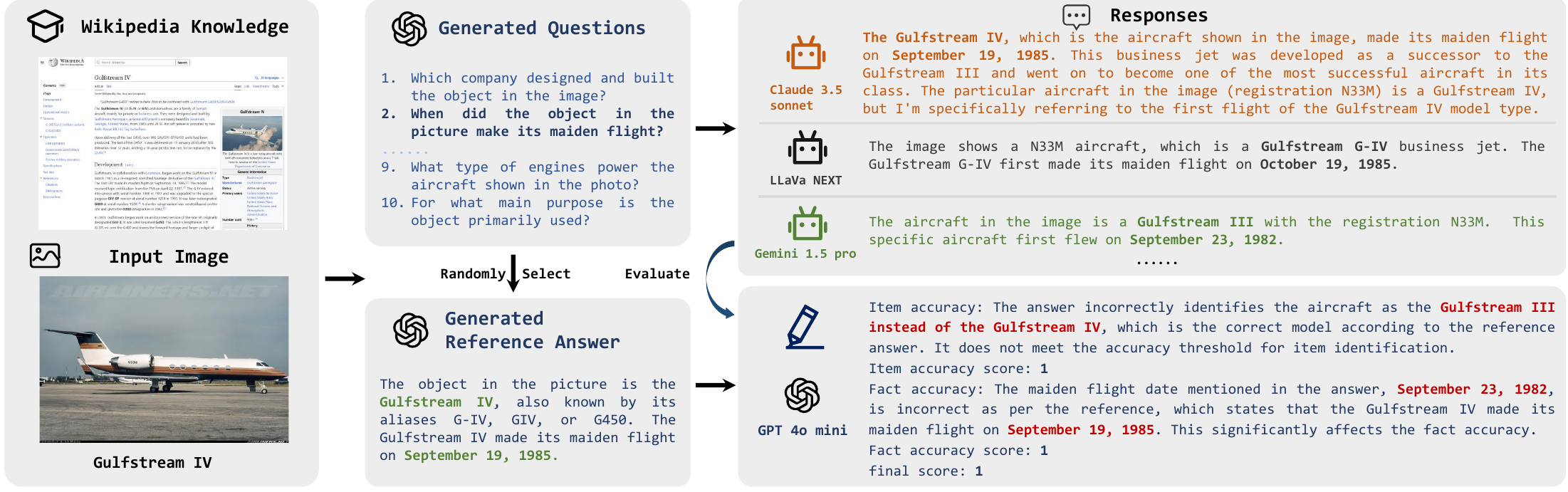}
    \caption{The proposed pipeline of data curation and evaluation examples.}
    \label{fig:pipeline}
    \vspace{-1.3em}
\end{figure*}

On the data front, we create \textit{mosaic} datasets by combining resized images into composite visuals, accompanied by specially crafted QA pairs tailored to this format. This approach boosts recognition accuracy by an average of 1\%. Building on this, we incorporate \textit{open-world} data featuring both introduction-style and open-ended questions. The introduction-style questions provide contextual background on image objects, with answers synthesized from Wikipedia entries using GPT-4o. In parallel, the open-ended QA follows a reference format similar to that depicted in Figure~\ref{fig:pipeline}. Together, these enhancements significantly outperform the baseline model, achieving at least a 10\% increase in recognition accuracy and a 6\%–12\% improvement in content accuracy across various categories in the FROW benchmark.

In terms of training optimization, we incorporate all fine-grained recognition data during the pretraining alignment phase and only a small portion during the Supervised Fine-Tuning (SFT) stage. This approach not only boosts the model's maximum fine-grained category recognition rate by up to 10\%, but also minimize the impact on other general capabilities. In summary, our contributions are summarized as follows:

\begin{itemize}[leftmargin=*, itemsep=0pt, topsep=0pt]

\item We introduce a more challenging fine-grained recognition benchmark and establish clear evaluation criteria. We then employ GPT-4o to evaluate the resulting responses. To the best of our knowledge, this is the first open benchmark specifically designed to evaluate fine-grained recognition capabilities.

\item Mosaic data and open-world data are constructed to significantly enhance the recognition accuracy and generalization of models on fine-grained recognition tasks.

\item By optimizing the data allocation at different stages of the training process, we not only improve the upper limit of the recognition ability of the model but also reduce its impact on general capabilities.

\end{itemize}

\section{Related work}

\subsection{Training stages of LVLMs}

Visual language models are typically trained in two main stages: pretraining for feature alignment and supervised fine-tuning. In the pretraining stage, models such as LLaVA \cite{liu2023llava} align visual inputs with textual descriptions by optimizing the projection matrix that connects these two modalities. During this process, the visual encoder and LLM weights remain frozen, ensuring that the model focuses solely on learning how to map visual features to the corresponding textual representations. Qwen-VL-chat \cite{Qwen-VL} similarly employs a large, weakly labeled dataset to achieve this alignment, enabling robust generalization across diverse visual and textual inputs. InternVL \cite{internvl} further enhances pretraining by incorporating multiple tasks, including image captioning, object detection and OCR, thereby fostering a more nuanced understanding of visual content. Furthermore, expanding the pretraining dataset to encompass more knowledge-rich data can significantly enhance model performance, particularly for fine-grained recognition tasks that require deeper contextual comprehension and precision.

\subsection{Benchmarks for evaluating LVLMs}
Understanding and reasoning are key capabilities for LVLMs. Understanding encompasses visual perception (evaluated by benchmarks like Q-Bench \cite{wu2024qbench} for tasks such as object detection) and contextual comprehension (e.g., MMNeedle \cite{wang2024multimodal} for long-context multimodal sequences), with modality-specific understanding like VideoNIAH \cite{zhao2024videoniah} focusing on spatio-temporal perception in video. Reasoning, on the other hand, assesses a model’s ability to integrate visual and domain-specific knowledge, such as Math-Vision \cite{wang2024measuring} for math-related tasks. While many benchmarks assess general capabilities, fine-grained recognition has often been neglected, and existing objective benchmarks \cite{geigle-etal-2024-african, VLMClassifier} fail to fully capture this aspect, highlighting the need for a subjective benchmark specifically targeting fine-grained recognition.

\subsection{Fine-grained recognition tasks for LVLMs}
To address the limitations in evaluating LVLMs on fine-grained object classification tasks, such as FOCI \cite{geigle-etal-2024-african} and ImageWikiQA \cite{VLMClassifier}. the FOCI benchmark systematically assesses LVLMs using domain-specific data subsets and negative sample mining in a multiple-choice format. Experimental results show that while LVLMs excel in image understanding and reasoning, their fine-grained classification performance remains inferior to CLIP models due to insufficient alignment between the image encoder and language model, highlighting the need for pretraining with finely annotated datasets. Additionally, ImageWikiQA \cite{VLMClassifier} finds that critical classification information exists in the latent space of VLMs but requires sufficient training data for effective decoding. A strong correlation between class exposure frequency and classification performance suggests that incorporating classification-relevant datasets during VLM training significantly enhances accuracy. However, ImageWikiQA’s multiple-choice format remains a constraint, as higher accuracy in such tasks does not necessarily translate to better recognition in open-world scenarios.

\begin{figure*}[t]
    \centering
    \includegraphics[width=\linewidth]{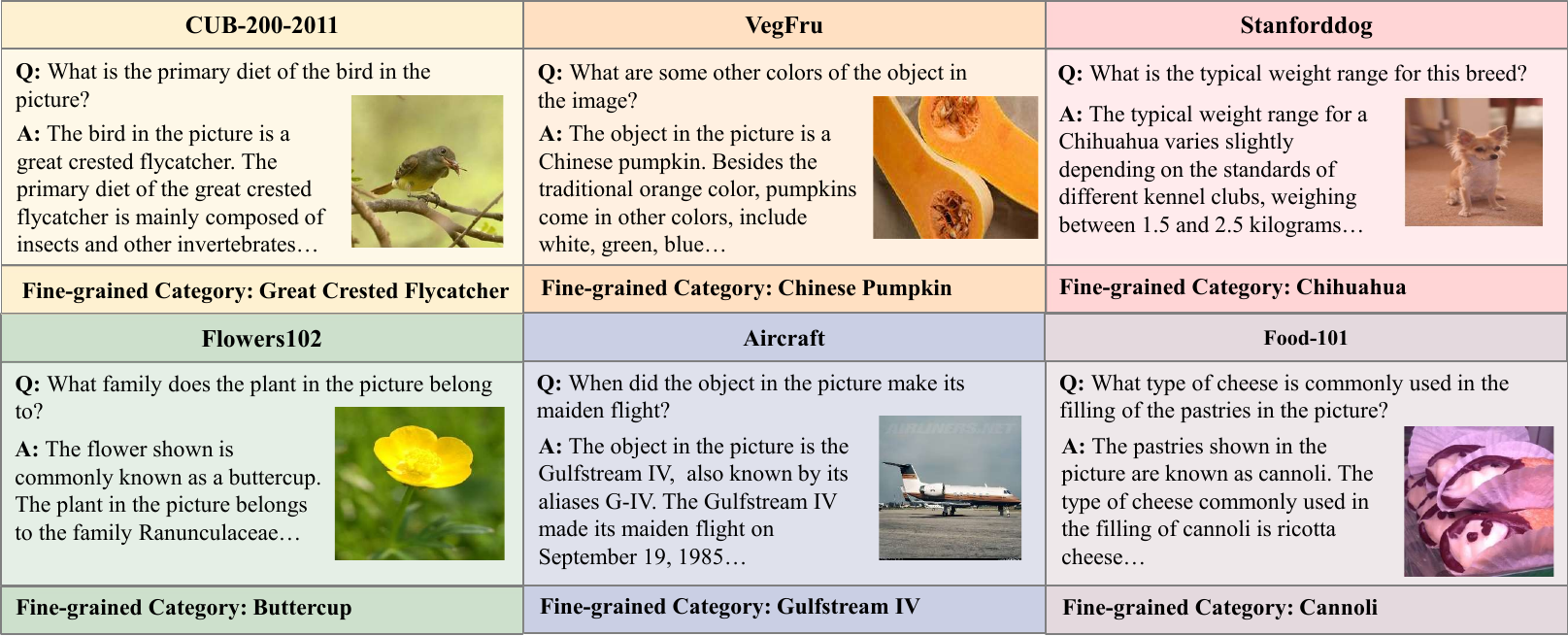}
    % \vspace{-10pt}
    \caption{Some examples of the benchmark, including questions and reference answers.}
    \label{fig:samples}
    \vspace{-1.3em}
\end{figure*}

\section{Fine-grained recognition open-world benchmark}
\label{sec:method_bk}
% Numerous benchmarks have been developed to evaluate the general capabilities of LVLMs. However, limited attention has been devoted to fine-grained tasks. While FOCI \cite{geigle-etal-2024-african} constructs multiple-choice questions based on fine-grained categories and ImageWikiQA \cite{VLMClassifier} introduces more complex queries, both remain restricted to multiple-choice formats. Such formats inherently simplify the task by narrowing the search space for the LVLM, often leading to higher accuracy. Consequently, these evaluations offer limited insight into the models’ performance on fine-grained recognition tasks. Therefore, developing a benchmark that employs open-ended questions is essential for a more comprehensive assessment of LVLM capabilities in fine-grained scenarios.

In this section, we will focus on the data curation and evaluation approach of the FROW benchmark, as well as the experimental results.

\subsection{Data curation}

To ensure both the quality and diversity of the dataset, we employ images sourced from six well-established fine-grained datasets: FGVC-Aircraft \cite{aircraft}, Caltech-UCSD Birds-200-2011 \cite{CUB}, Food-101 \cite{food101}, Stanford Dogs \cite{stanforddog}, Oxford Flowers-102 \cite{nilsback2008automated}, and VegFru \cite{hou2017vegfru}. A total of 859 categories are extracted from these datasets. For each category, a single image is selected from the respective testset, and then a question is generated using GPT-4o, based on category information obtained from Wikipedia.

The data curation pipeline is illustrated in Figure~\ref{fig:pipeline}. First, for each fine-grained category, we retrieve relevant information from Wikipedia and subsequently prompt GPT-4o to generate questions based on the retrieved content. We ensure that the generated questions remain pertinent without explicitly mentioning the category itself, thereby requiring the model to accurately infer the fine-grained category prior to answering. Next, we prompt GPT-4o to filter out both low-quality and unanswerable questions, and finally, we manually verify the validity of the remaining questions.

To facilitate subsequent evaluation of LVLM’s responses, we then prompt GPT-4o to produce a referenced answer grounded in both the fine-grained category of the object and the relevant Wikipedia information. Because objects may have multiple valid names for the same fine-grained category, we instruct GPT-4o to include all potentially correct category names in its referenced answer. The detailed representations of the samples in FROW are provided in Figure \ref{fig:samples}.

\subsection{Evaluation metrics and results}

\paragraph{Evaluation model}

To evaluate the quality of open-ended responses generated by LVLMs on our benchmark, we adopt the approach of employing advanced proprietary models as evaluators, following recent studies \cite{xu2023superclue, liu2023g}. Specifically, we initially consider GPT-4o and GPT-4o-mini, two widely recognized and sophisticated proprietary models, for this task. Through extensive experimentation, we observe that both models produce consistent and accurate scores when provided with well-defined evaluation criteria. Considering the observed consistency and the necessity of cost efficiency, we ultimately select GPT-4o-mini as the primary evaluator for assessing responses.

Prior to deciding on GPT-4o-mini for evaluation, we conduct a comparative study involving human annotators, GPT-4o, GPT-4o-mini and Claude 3.5 sonnet. We score 300 randomly sampled LVLM responses and find that GPT-4o and GPT-4o-mini produce nearly identical results, with an average discrepancy of less than 1\%. Additionally, the discrepancy between Claude3.5 and GPT-4o is within 1.1\%. The discrepancy between human annotations and GPT-4o is within 1\%. 

\begin{table}[!t]
\centering
\caption{The following results pertain to the evaluation of various LVLM models on the FROW benchmark. An asterisk (*) denotes outcomes obtained after applying the optimization strategy. The optimized versions of InternVL and LLaVA demonstrate performance gains exceeding 10 points, with InternVL achieving results that closely approximate the performance of GPT-4o.}
\resizebox{\textwidth}{!}{%
\begin{tabular}{llcccccc}
\toprule
& Model & CUB-200-2011 & Stanford Dog & Flowers102 & Food101 & VegFru & Aircraft \\
\midrule
\multirow{3}{*}{Proprietary} 
  & GPT-4o \cite{hurst2024gpt}     & 64.20 & 62.83 & 70.98 & 75.00 & 60.17 & 63.80 \\
  & Claude 3.5 sonnet \cite{anthropic2024claude}  & 59.30 & 62.00 & 73.53 & 68.60 & 54.94 & 64.40 \\
  & Gemini 1.5 pro \cite{team2024gemini} & 62.40 & 69.33 & 65.10 & 69.20 & 63.29 & 50.80 \\
  & GLM v plus \cite{glm2024chatglm} & 49.90 & 52.33 & 56.67 & 62.60 & 43.04 & 42.60 \\
  & Doubao 1.5 vision pro & 48.10 & 59.33 & 64.51 & 70.60 & 54.60 & 49.40 \\
  & Step 1v & 51.30 & 62.33 & 59.22 & 70.40 & 50.30 & 51.80 \\
  & Hunyuan vision \cite{sun2024hunyuan} & 34.10 & 40.50 & 46.67 & 58.60 & 42.45 & 29.60 \\
\cmidrule(r){1-8}
\multirow{5}{*}{Open Source} 
  & Qwen-VL-chat-78B \cite{Qwen-VL}     & 28.80 & 48.67 & 53.73 & 57.20 & 42.87 & 35.80 \\
  & InternVL 2.5-8B \cite{internvl}    & 27.70 & 24.92 & 30.98 & 42.53 & 21.52 & 16.80 \\
  & LLaVA 1.5-7B \cite{liu2023llava}    & 15.80 & 19.50 & 19.02 & 39.00 & 21.60 & 16.40 \\
  \cmidrule(r){2-8}
  & InternVL 2.5-8B*   & 54.70 (+27.00) & 57.17 (+32.25) & 58.43 (+27.55) & 63.00 (+20.63) & 55.27 (+33.75) & 31.80 (+15.00) \\
  & LLaVA 1.5-7B*   & 28.70 (+12.90) & 53.00 (+33.50) & 29.80 (+10.78) & 49.40 (+10.40)  & 40.25 (+18.55) & 25.60 (+9.20) \\
\bottomrule
\end{tabular}%
}
\label{tab:scores}
\vspace{-7pt}
\end{table}

\paragraph{Evaluation criteria}

When designing the evaluation criteria, we comprehensively assess the model’s answers from two perspectives: 1) \emph{recognition accuracy} and 2) \emph{content accuracy}.

Specifically, since the questions in the benchmark are constructed based on fine-grained category information of objects, the recognition accuracy evaluates whether the model can identify the fine-grained category of the object. Considering the model's response to category may lack precision, we define a three-level evaluating system, ranging from 0 to 2 points, to reflect the precision of category recognition. Higher scores are awarded for more precise recognition. For example, consider a scenario in which the ground truth label is ``Boeing 737-600”. Suppose two responses are provided: ``Boeing 737” and ``Boeing 737-200”. In this case, the first response would receive one point, while the second response would receive zero points. To ensure the scoring is well-founded, we will first have GPT-4o-mini explain the rationale behind its evaluation, and then assign an appropriate score. 

The content accuracy is assessed by evaluating the correctness of the information provided for fine-grained categories within the response. To achieve this, we first prompt GPT-4o to generate a referenced answer using the name of the object and relevant information sourced from Wikipedia. Subsequently, we define a four-level evaluation system, based on the referenced answer, to assess the correctness of the model's response. In this system, a score of 0 indicates that the response contains no correct information. Scores of 1 or 2 reflect the presence of partial errors, with the specific score determined by GPT-4o-mini. A score of 3 indicates that the response is completely accurate. The detailed evaluation prompt is outlined in Appendix \ref{app:evaluation_metrics}.

Finally, the average values of the recognition accuracy score and the content accuracy score are utilized as the final evaluation metrics.

\paragraph{Evaluation results}
We assess the performance of current open-source and proprietary large vision-language models (LVLMs) using our benchmark, with scores converted to a percentage scale as detailed in Table \ref{tab:scores}. The proprietary models are evaluated include GPT-4o \cite{hurst2024gpt}, Claude-3.5-Sonnet \cite{anthropic2024claude}, and Gemini-1.5-Pro \cite{team2024gemini}, while the open-source models comprise Qwen-VL-chat \cite{Qwen-VL}, InternVL-2.5 \cite{internvl}, and LLaVA-1.5 \cite{liu2023llava}. 

Table \ref{tab:scores} demonstrates that current open-source models continue to underperform on FROW benchmarks. Notably, even larger models, such as Qwen-VL-Chat, exhibit a significant performance gap compared to GPT-4o. To address this issue, we select the open-source models InternVL 2.5, both with 8 billion parameters, to explore strategies for data construction and training optimization.

\section{Optimization strategies for fine-grained recognition}
\label{sec:method_op}
In Section \ref{sec:conf}, we will provide an overview of the models and datasets utilized in our study. In Section \ref{sec:data}, we will focus on the data perspective, discussing what types of data can enhance the model's fine-grained recognition capabilities. In Section \ref{sec:train}, we will explore strategies from the training stage perspective to effectively enable the model to learn fine-grained knowledge.

\subsection{Experimental setup} \label{sec:conf}
\paragraph{Model} Considering the training costs, we perform experimental validation on InternVL-2.5-8B~\cite{internvl}. The vision encoder and language model parameters are initialized using their respective original weights.

\paragraph{Dataset} 
For the training data, we incorporate both general data, consisting of 558K aligned samples and 665K supervised fine-tuning samples from LLaVA \cite{liu2023llava}, and fine-grained data, which consist of open-ended questions generated based on the training sets of six well-established fine-grained datasets. For the validation data, we evaluate the LVLM's general capabilities on several popular benchmarks, including AI2D \cite{kembhavi2016diagram}, ChartQA \cite{ChartQA}, DocVQA \cite{DocVQA}, InfographicsVQA \cite{mathew2022infographicvqa}, MathVista \cite{lu2024mathvista}, and POPE \cite{li2023evaluating}. To assess fine-grained capabilities, we use our fine-grained benchmark and open-ended answers generated from the test sets of the fine-grained datasets. Details of the generated fine-grained training and validation samples are provided in Appendix~\ref{app:fg_train_val_data}.

\begin{figure*}[t]
    \centering
    \includegraphics[width=\linewidth]{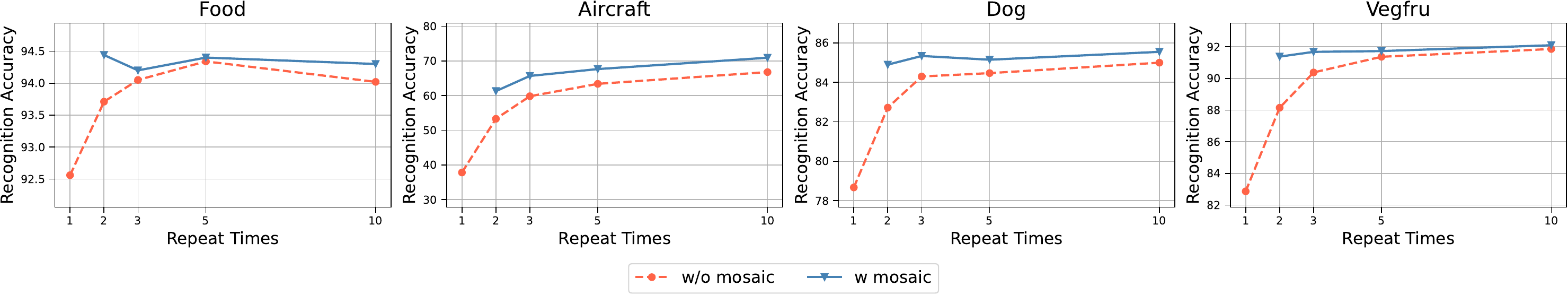}
    \vspace{-1.5em}
    \caption{Comparison of recognition accuracy among different repetitions of dataset. As the number of repetitions in the dataset increases, the recognition rate shows signs of convergence. However, the model with the inclusion of mosaic data converges faster and achieves a higher accuracy upon convergence.}
    \label{fig:comparison2}
    % \vspace{-1em}
\end{figure*} 

\subsection{Data construction}
\label{sec:data}
\begin{wrapfigure}{r}{0.55\textwidth}
    \centering
    \vspace{-13pt}
    \includegraphics[width=0.99\linewidth]{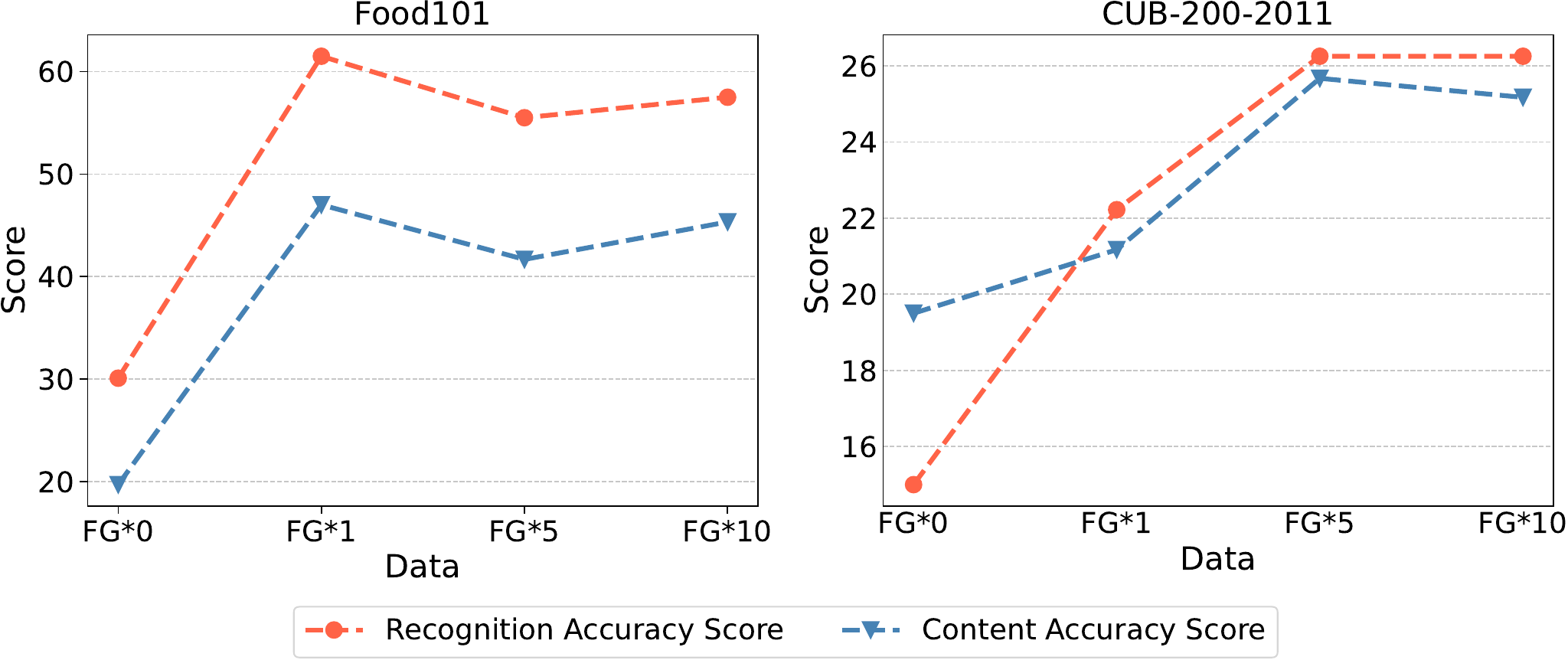}
    % \vspace{-25pt}
    \caption{``FG*n'' denotes that the fine-grained data are repeated n times during training. Each experiment contributes to the overall 558K SFT dataset. The reported values represent the scores from the FROW benchmark.}
    % \vspace{-1em}
    \label{fig:diverse}

    \includegraphics[width=0.99\linewidth]{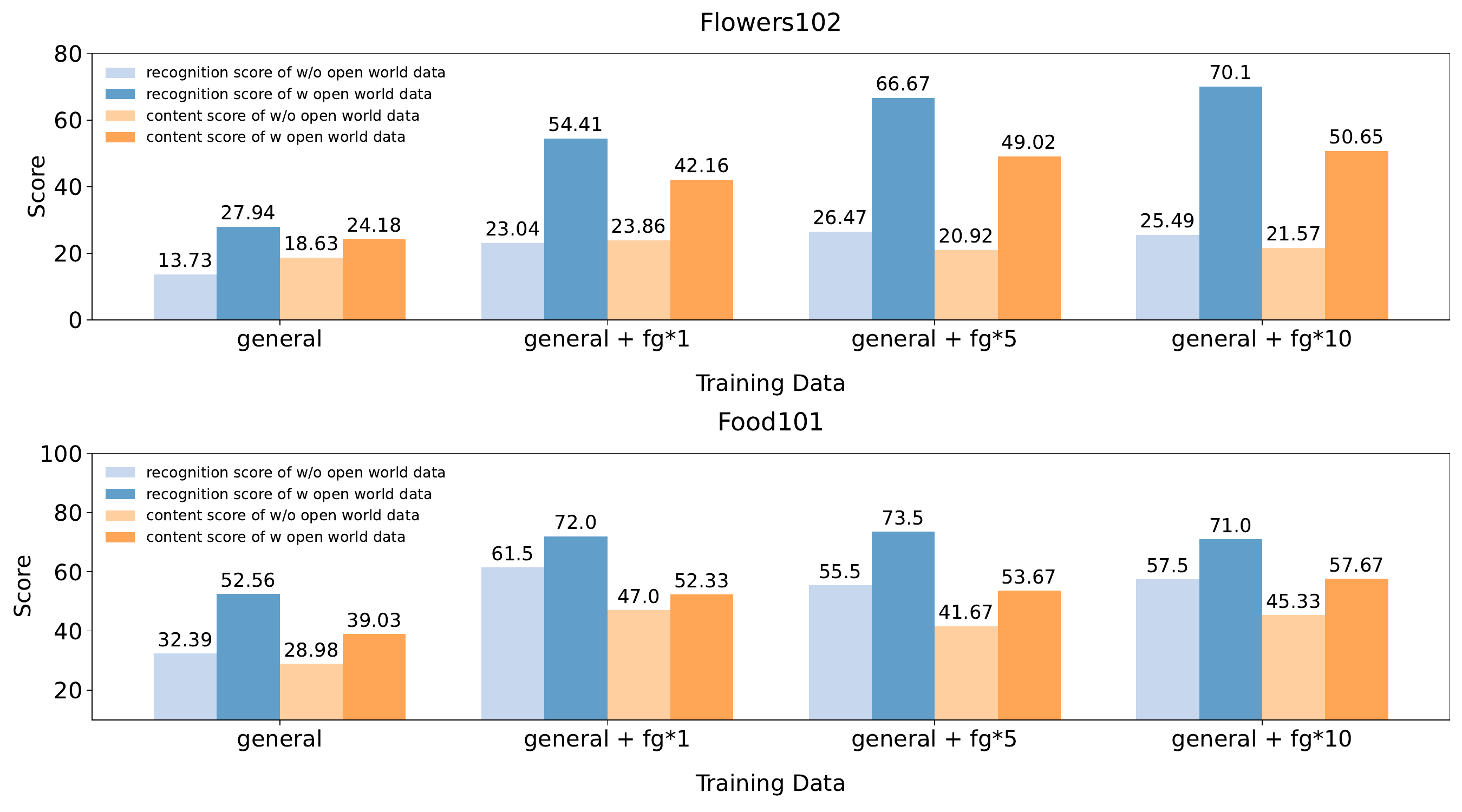}
    % \vspace{-25pt}
    \caption{The general data is 665K SFT data and ``FG*n'' represents the fine-grained data are repeated n times during training. The dark-colored bars represent the scores when open world data is included, while the light-colored bars indicate the scores without open world data. Red represents the recognition accuracy scores, and blue represents the content accuracy scores.}
    \vspace{-2em}
    \label{fig:comparison_diverse}
\end{wrapfigure}

In this section, we validate the effectiveness of the proposed data during the SFT stage. The recognition accuracy is assessed using the short-answer benchmark, as the model is trained on short-answer data. Notably, this data also enhances performance on the \textbf{FROW} benchmark, as illustrated in Figure \ref{fig:comparison_diverse}.

\paragraph{Mosaic data}

\begin{wrapfigure}{r}{0.5\textwidth}
  \centering
    \vspace{-12pt}
    \includegraphics[width=0.99\linewidth]{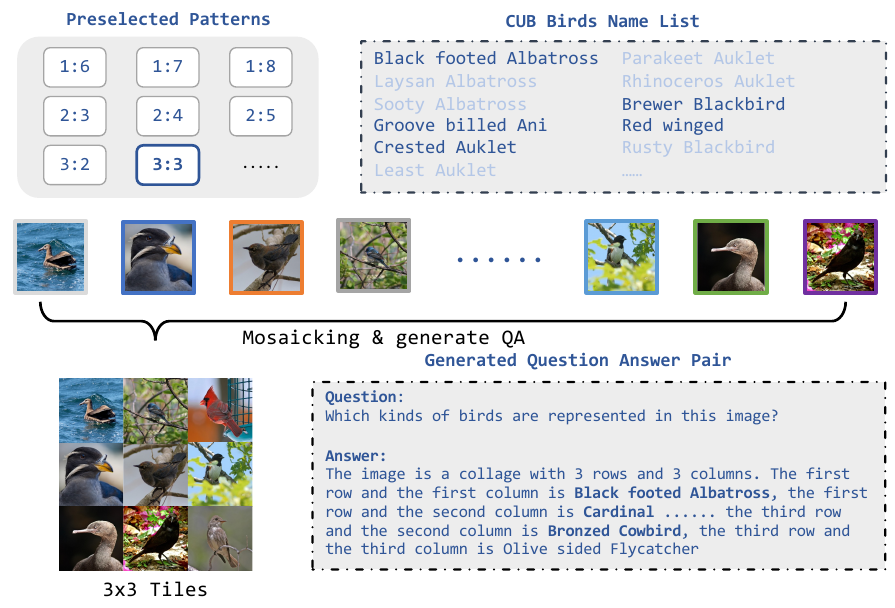}
    \vspace{-14pt}
    \caption{The process of constructing mosaic images: First, a pattern is randomly selected. Based on the chosen pattern, the same number of categories are randomly selected, and from each category, one image is randomly picked. These images are then combined to form a \(m \times n\) mosaic image. The example shown here illustrates a \(3 \times 3\) pattern.}
    \vspace{-1em}
    \label{fig:mosaic}
\end{wrapfigure}

During our experiments, we observe that fine-grained recognition tasks in LVLM require data to be repeated many times to reach convergence, nearly 10 times, and the performance ceiling is relatively low. This leads us to consider whether data augmentation in recognition tasks could be effective. We experiment with using AutoAugment\cite{autoaugment} for data augmentation during training and it really improves the performance, but it cannot reduce the amount of data. To address this, we propose an image mosaic strategy, where images of the same class are mosaicked together following a fixed pattern. We then construct question-answer pairs based on the mosaic images. The way to construct mosaic data is shown in Figure \ref{fig:mosaic}. We find that this method not only improves recognition accuracy, but also greatly reduces the amount of training data required. 

To facilitate comparison, we first establish a specific pattern for the mosaic image, using a \(3 \times 3\) format in the experiment. Since each mosaic image consists of 9 individual images, it is repeated 9 times to match the amount of data of the short answer dataset. As shown in Figure \ref{fig:comparison2}, under the same amount of data, the model converges more quickly when using mosaic images and achieves higher recognition accuracy. This suggests that mosaic images are particularly effective.

\begin{figure}[!t]
    \centering
    \includegraphics[width=\linewidth]{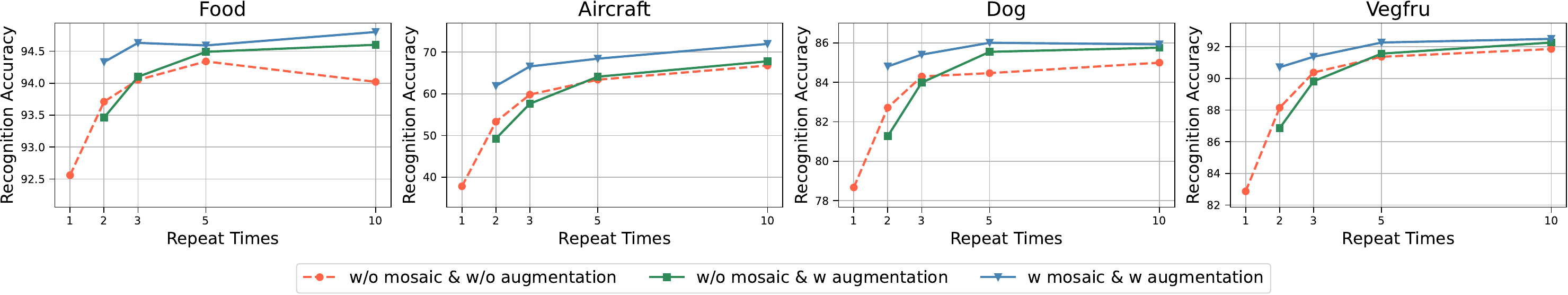}
    \vspace{-1.5em}
    \caption{Comparison between the two ways of data augmentation. The blue represents the recognition accuracy without any data augmentation or mosaic data, the green represents the accuracy with only traditional data augmentation, and the orange represents the accuracy with both data augmentation and the addition of mosaic data.}
    \label{fig:aug}
    \vspace{-1em}
\end{figure}

Meanwhile, this kind of data augmentation doesn't conflict with the traditional data augment, it could still improve the performance with AutoAugment. In the Figure \ref{fig:aug}, when the number of repetitions is low, models trained with two types of data augmentation may not have fully converged, which is why their performance is not significantly higher than models trained with only traditional data augmentation. However, as the number of repetitions increases, the models trained with these two augmentation methods gradually converge and consistently outperform others, achieving the highest recognition rate. 

\paragraph{Open world data}

Incorporating short-answer data and mosaic data can enhance the model's fine-grained recognition capabilities, leading to improved benchmark performance. 
However, exclusive reliance on these data types does not guarantee consistent performance gains and may result in stagnation or even a decline in content accuracy. The corresponding results are presented in Figure~\ref{fig:diverse}.

Therefore, we incorporate introduction-type data and open-ended data to inject additional knowledge into the model. At the same time, we remove questions from the training data that are identical to those in the benchmark to ensure that the model's benchmark performance more accurately reflects its true capabilities.

For data related to introductions, we predefine a set of questions designed to elicit background information about the object depicted in the image. To generate the corresponding introduction content, we prompt GPT-4 to summarize relevant information sourced from Wikipedia. To minimize potential inaccuracies in the generated summaries, we request GPT-4 to produce each summary three times. This process improves the reliability and robustness of the dataset for the introduction. The method for generating open-ended data follows the same procedure used to construct reference answers in benchmark datasets. Accordingly, for each category, the dataset consists of three introduction-type QA pairs and two open-ended QA pairs.

Figure \ref{fig:comparison_diverse} clearly demonstrates that incorporating open-world data significantly improves both recognition and content accuracy scores, with improvements ranging from 15\% to 45\%.

\subsection{Training process}\label{sec:train}

In the previous section, we discuss how fine-grained recognition capabilities of LVLMs can be enhanced from the perspective of data. In this section, we address the same issue from the perspective of the training process. For clarity, we break down the training into four stages. The first stage is the pretraining stage, where the visual and language models are pretrained separately, which is outside the scope of our discussion. The second stage is the alignment stage, where the alignment module is trained to bridge the gap between the vision and language modalities. The third stage is the fine-tuning stage, which aims to enhance the LVLM’s overall capabilities. Finally, the fourth stage focuses on improving the model's performance on specific tasks.

In the following, we first detail the limitations of directly fine-tuning LVLMs on fine-grained data during the third and fourth stages. We then present our findings that the alignment module in LVLMs can also learn fine-grained knowledge. Finally, we propose a training paradigm that enables LVLMs to retain their general capabilities while acquiring strong fine-grained recognition abilities.

\begin{table*}[!t]
    \centering
    \small
    \caption{Results of InternVL trained under different settings on fine-grained and general tasks. The ``558k'' represents the alignment data, ``665k'' represents the generic fine-tuneing data, while ''fg'' represents the fine-grained data used in training. ``Short Answer'' represents the results on questions about the object fine-grained category.}
    \label{table:fg_general_performance}
    \resizebox{\textwidth}{!}{
    \begin{tabular}{cclccccccc}
        \toprule
        \multirow{2}{*}{Setting} & \multicolumn{3}{c}{\rule{0pt}{9pt}Training Process} & \multicolumn{6}{c}{General Capabilities}                             \\
        \cmidrule(r){2-10}
                                 & \rule{0pt}{7pt}Aligment     & FT          & FT    & \emph{AI2D}   & \emph{ChartQA} & \emph{DocVQA} & \emph{InfographicsVQA} & \emph{MathVista} & \emph{POPE}  \\
        \cmidrule(r){1-10}
        \#1                      & \rule{0pt}{9pt}558k         & 665k        & --    & 65.31 & 27.36  & 42.43 & 30.27               & 22.6      & 87.6  \\
        \#2                      & 558k         & 665k        & fg    & 48.67 & 13.45  & 20.36 & 18.89               & 16.7      & 83.39 \\
        \#3                      & 558k         & 665k+fg     & --    & 65.02 & 26.79  & 41.11 & 28.34               & 23.2      & 87.6  \\
        \#4                      & 558k         & fg          & --    & --     & --      & --     & --                   & --        & --    \\
        \midrule
        \multirow{2}{*}{Setting} & \multicolumn{3}{c}{\rule{0pt}{7pt}Training Process} & \multicolumn{6}{c}{Fine-grained Recognition Capabilities - Short Answer}                        \\
        \cmidrule(r){2-10}
                                 & \rule{0pt}{9pt}Alignment     & FT          & FT    & \emph{Aircraft}    & \emph{CUB}     & \emph{Flowers102} & \emph{Food-101}     & \emph{Dog}  & \emph{VegFru} \\
        \cmidrule(r){1-10}
        \#1                      & \rule{0pt}{7pt}558k         & 665k        & --    & --     & --      & --     & --                   & --        & --    \\
        \#2                      & 558k         & 665k        & fg    & 68.4   & 83.32   & 92.66  & 94.03                & 84.51     & 91.65 \\
        \#3                      & 558k         & 665k+fg     & --    & 66.03  & 83.84   & 92.19  & 94.46                & 85.33     & 90.77 \\
        \#4                      & 558k         & fg          & --    & 69.45  & 83.43   & 93.54  & 94.25                & 84.41     & 91.79 \\
        \midrule
        \multirow{2}{*}{Setting} & \multicolumn{3}{c}{\rule{0pt}{9pt}Training Process} & \multicolumn{6}{c}{Fine-grained Capabilities - FROW}                        \\
        \cmidrule(r){2-10}
                                 & \rule{0pt}{7pt}Alignment     & FT          & FT    & \emph{Aircraft}    & \emph{CUB}     & \emph{Flowers102} & \emph{Food-101}     & \emph{Dog}  & \emph{VegFru} \\
        \cmidrule(r){1-10}
        \#1                      & \rule{0pt}{7pt}558k         & 665k        & --    & --     & --      & --     & --                   & --        & --    \\
        \#2                      & 558k         & 665k        & fg    & 38.17  & 54.75   & 61.69  & 76.75                & 59.79     & 62.66 \\
        \#3                      & 558k         & 665k+fg     & --    & 41.34  & 55.38   & 56.36  & 64.25                & 56.88     & 59.78 \\
        \#4                      & 558k         & fg          & --    & 44.17  & 55.38   & 59.40  & 62.25                & 57.22     & 59.32 \\
        \bottomrule
    \end{tabular}
    }
    \vspace{1em}
\end{table*}

\paragraph{Evaluating the viability of the fourth stage}
% \paragraph{Fine-tuning LVLMs on fine-grained data during the fourth stage leads to weak general capabilities.}
As shown in Table~\ref{tab:scores}, the fine-grained recognition capabilities of LVLMs are limited. A straightforward approach to improve fine-grained performance is to fine-tune LVLMs on fine-grained data (detailed in Appendix~\ref{app:fg_train_val_data}). However, as demonstrated in Table~\ref{table:fg_general_performance}, while the fine-tuned model shows improvements in fine-grained recognition, its performance on general tasks drops significantly compared to the original model (\#1 and \#2). This indicates that directly fine-tuning LVLMs on fine-grained data is not a viable solution.

% \paragraph{There is a trade-off between imporving fine-grained and general capabilities.}
\paragraph{Evaluating the viability of the third stage}
Previous results indicate that directly fine-tuning LVLMs on fine-grained data leads to significant forgetting of general knowledge. To address this, we experimented with merging fine-grained and general data during the fine-tuning process. As shown in Table~\ref{table:fg_general_performance}, the model trained on mixed data (\#3) demonstrates promising performance on both fine-grained and general tasks. However, its performance on general tasks is weaker compared to the model trained solely on general data (\#1 and \#3). Similarly, its performance on fine-grained tasks is also lower than that of the model trained exclusively on fine-grained data (\#3 and \#4). These findings suggest that the trade-off between learning fine-grained and general knowledge requires careful consideration, which is also mentioned by MM1 \cite{MM1}).

\begin{figure*}[t]
    \centering
    {\includegraphics[width=0.99\textwidth]{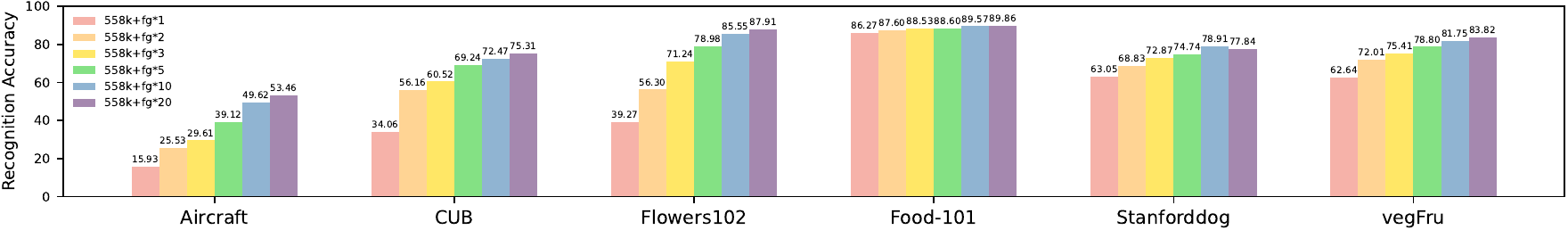}}
    \caption{Fine-grained rcognition accuracy of LVLMs trained on alignment data, with 558k general data and fine-grained data. ``fg*n'' represents the fine-grained data are repeated n times during training.}
    \label{fig:st2_fg}
\end{figure*}

\paragraph{Evaluating the viability of stage 2}

\begin{figure*}[t!]
    \centering
    {\includegraphics[width=0.99\textwidth]{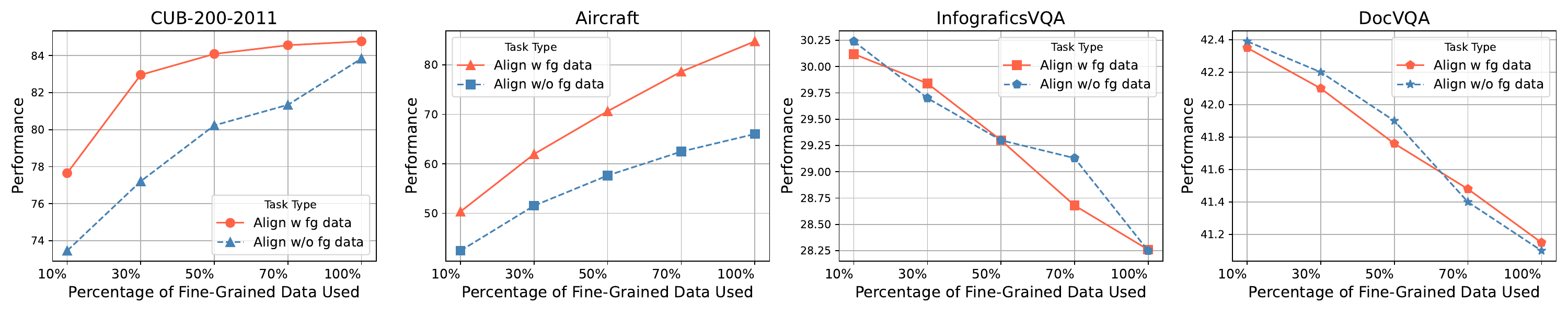}}
    % \vspace{-1em}
    \caption{Performance of LVLMs on general and fine-grained tasks after fine-tuning with varying proportions of fine-grained data. The origin point represents models trained with fine-grained data included in the alignment stage, while blue dots indicate models trained without fine-grained data in the alignment stage.}
    \label{fig:mixed_tasks_results}
    % \vspace{-15pt}
\end{figure*}

To address the trade-off between fine-grained and general data, we experiment with incorporating fine-grained data at different stages to minimize their conflicting effects. The alignment training stage typically uses easily accessible, low-information image-text pairs to train the alignment module, with the goal of bridging the gap between vision and language modalities, rather than enhancing the model's capabilities. However, when fine-grained data is incorporated into the alignment data, as shown in Figure~\ref{fig:st2_fg}, the alignment module demonstrate considerable fine-grained recognition ability when the amount of fine-grained data is sufficiently large, which suggests the feasibility of incorporating fine-grained data into the alignment process of training to enhance the LVLM's fine-grained recognition capabilities.

% \paragraph{ Incorporating fine-grained data during the alignment stage brings benefit for fine-grained capabilities, without harming general capabilities.}
\paragraph{Data allocation}
Building on the above findings, we propose incorporating fine-grained data during the alignment training stage to improve the fine-grained recognition capabilities of LVLMs. As shown in Figure~\ref{fig:mixed_tasks_results}, LVLM trained with fine-grained data during the alignment stage achieve strong fine-grained performance with only a small amount of fine-grained data in the fine-tuning stage. With the same amount of data, its fine-grained recognition accuracy consistently outperforms those do not include fine-grained data in the alignment stage. Regarding general capabilities, as illustrated in Figure~\ref{fig:mixed_tasks_results}, the model trained with fine-grained data during the alignment stage maintains comparable performance to models without fine-grained data, demonstrating that our proposed training paradigm can significantly enhance the fine-grained capabilities of LVLMs while preserving their general abilities. Additional results can be found in Appendix~\ref{app:more_res}.

% \section{Limitations}
% The impact of post-training stages in large models, including preference alignment via methods such as RLHF or DPO, on fine-grained recognition accuracy remains underexplored. Despite its importance, limited research has examined how these alignment techniques influence detailed recognition tasks. Currently, our model training is confined to the SFT stage. Future studies should prioritize investigating the later stages of training to unlock their potential and optimize the design of training datasets, thereby enhancing overall model performance.

% Our current exploration of mosaic data remains relatively superficial, lacking a comprehensive theoretical understanding of its full potential. Although its utility has been demonstrated in certain tasks, the intrinsic value and overall effectiveness of this data format have not been rigorously evaluated. To substantiate its usefulness, additional experiments are required to provide more definitive evidence of its impact. Moreover, diversifying the methods for constructing mosaic data could significantly enhance its utility by generating more robust datasets. A broader and more innovative approach to data construction is likely to improve model training outcomes by offering richer and more varied inputs. This, in turn, may lead to better model alignment and improved fine-grained recognition in subsequent stages of development.

\section{Conclusion}
% Our paper focuses on the fine-grained recognition abilities of LVLM, which have expanded beyond text inputs to include video, images, and audio. While there has been a rapid development of various benchmarks designed to evaluate LVLM performance from different angles, many of these benchmarks focus on high-level tasks that are often complex and difficult. However, they tend to overlook a crucial aspect: the fine-grained recognition capabilities of LVLMs. This area remains relatively underexplored in current evaluations, and existing benchmarks do not provide a sufficient way to assess these recognition abilities in detail.

% To address this gap, we propose a new fine-grained recognition benchmark, rated by GPT-4o, designed specifically to evaluate LVLM performance on more subtle recognition tasks. In addition to this benchmark, we explore ways to improve the recognition capabilities of open-source LVLMs through targeted data construction and training strategies. Our experiments show that it is possible to enhance fine-grained recognition performance without compromising the model's general capabilities.

Our work focuses on evaluating and enhancing the fine-grained recognition capabilities of LVLMs, a largely overlooked aspect in existing research. To assess these capabilities, we introduce FROW, a fine-grained benchmark that employs open-ended questions to evaluate LVLMs on complex recognition tasks. Additionally, we propose strategies in data construction and training processes to improve fine-grained recognition while preserving general task performance.

\section*{Limitations} 
While our training strategy enhances LVLMs' fine-grained recognition, methods like preference alignment via RLHF or DPO remain underexplored, as our training is currently limited to the SFT stage. In data construction, our exploration of mosaic data is still preliminary, requiring further experiments to validate its effectiveness.
\bibliographystyle{unsrt}
\bibliography{reference}
% \bibliographystyle{NeurIPS2025}

%%%%%%%%%%%%%%%%%%%%%%%%%%%%%%%%%%%%%%%%%%%%%%%%%%%%%%%%%%%%
\newpage
\appendix
\section{Benchmark}\label{app:benchmark}

\subsection{Dataset}\label{app:dataset}
We source images from six fine-grained datasets. Table~\ref{table:fg-datasets} indicates their meta-classes, the amount of samples, the number of categories. For all datasets, we construct human-oriented evaluation questions based on their test sets. We use the original labels directly from the datasets for the machine-oriented evaluation.

\begin{table}[h]
    \caption{Details of six fine-grained datasets sorted by their numbers of categories. ``Meta-class'' refers to a high-level categorization of the dataset. ``Categories'' refers to the number of fine-grained categories. ``Samples'' refers to the total number of samples in each dataset.}
    \label{table:fg-datasets}
    \centering
    \begin{tabular}{lccc}
        \toprule
        Datasets                     & Meta-class  & Categories & Samples \\
        \midrule
        \emph{FGVC Aircraft \cite{aircraft}} & Aircraft    & 100    & 6,667       \\
        \emph{Food101 \cite{food101}}       & Food        & 101    & 101,000     \\
        \emph{Flowers102 \cite{nilsback2008automated}}    & Flower      & 102    & 7,169       \\
        \emph{Stanford Dogs \cite{stanforddog}} & Dog         & 120    & 20,580      \\
        \emph{CUB-200-2011 \cite{CUB}}  & Bird        & 200    & 11,788      \\
        \emph{VegFru \cite{hou2017vegfru}}        & Vegetable   & 292    & 146,131     \\
        \bottomrule
    \end{tabular}
\end{table}

\subsection{Prompts}\label{app:evaluation_metrics}
% 1.recognition score-The detailed evaluating criteria are provided in Appendix~\ref{}. 
% 2.factual score-The evaluating criteria are detailed in Appendix~\ref{}. 
% 3.As shown in Figure~\ref{fig:scores}, the accuracy of LVLMs in fine-grained category recognition is significantly lower than that of fine-grained tailored models, which generally achieve accuracy over 90\% on fine-grained datasets
% 4.Results of InternVL-2.5 and LLaVA-1.5 on fine-grained multiple-choice question. The question creation is detailed in Appendix~\ref{app:evaluation_metrics}
\begin{figure}[hbtp]
    \centering
    \includegraphics[width=0.99\linewidth]{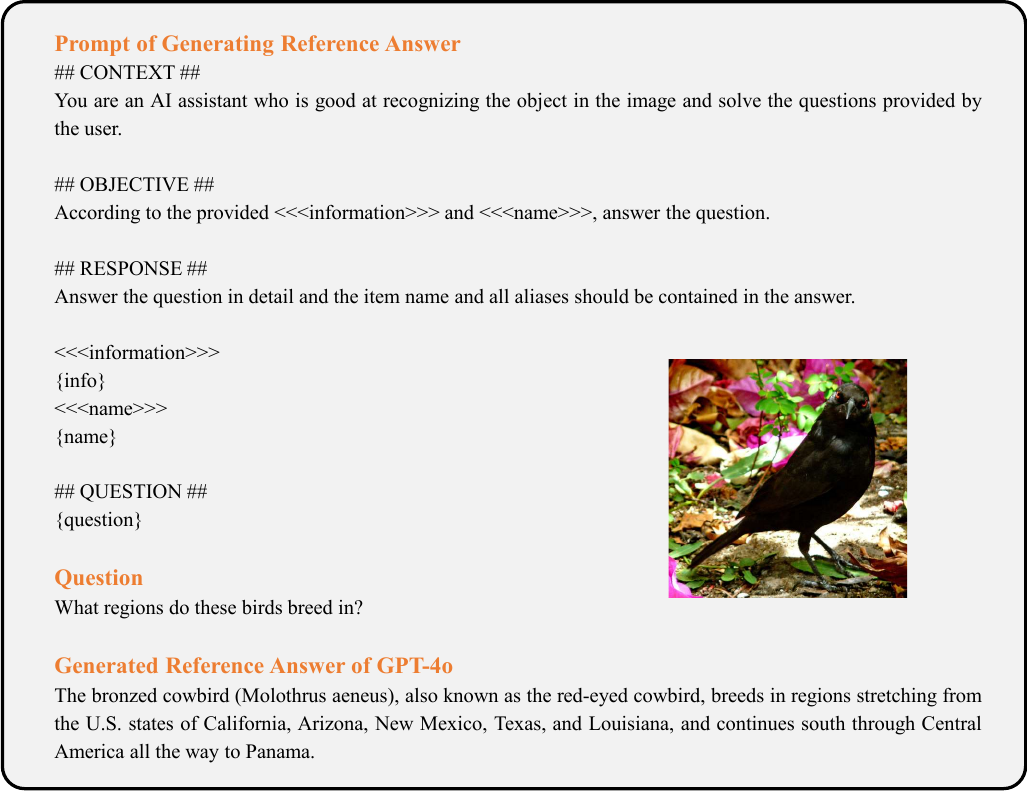}
    \caption{Prompt of generating reference answer.}
    \label{fig:prompt2}
\end{figure}
\begin{figure}[htbp]
    \centering
    \includegraphics[width=0.8\linewidth]{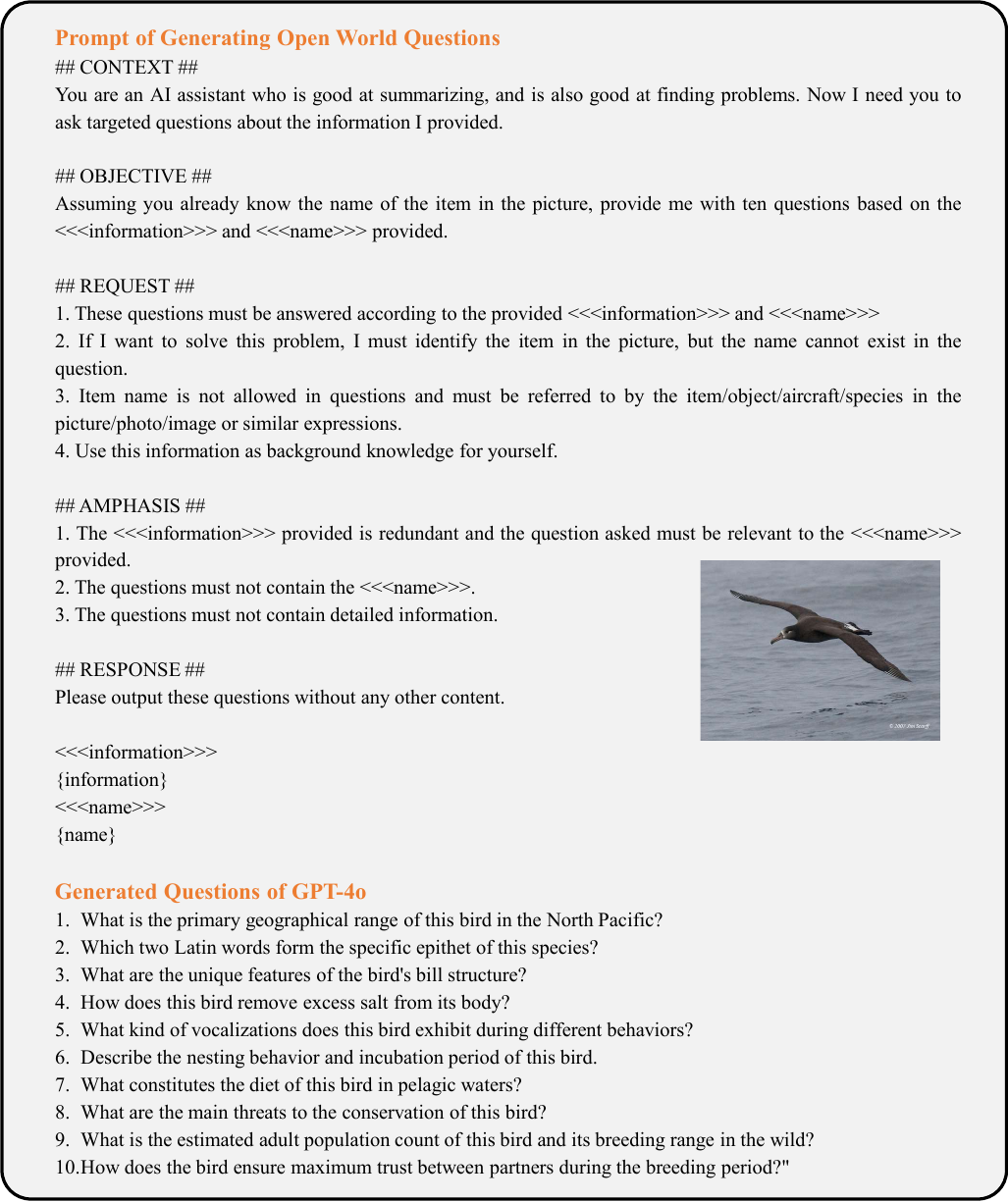}
    \caption{Prompt of generating open world questions.}
    \label{fig:prompt1}
\end{figure}
\begin{figure}[htbp]
    \centering
    \includegraphics[width=0.8\linewidth]{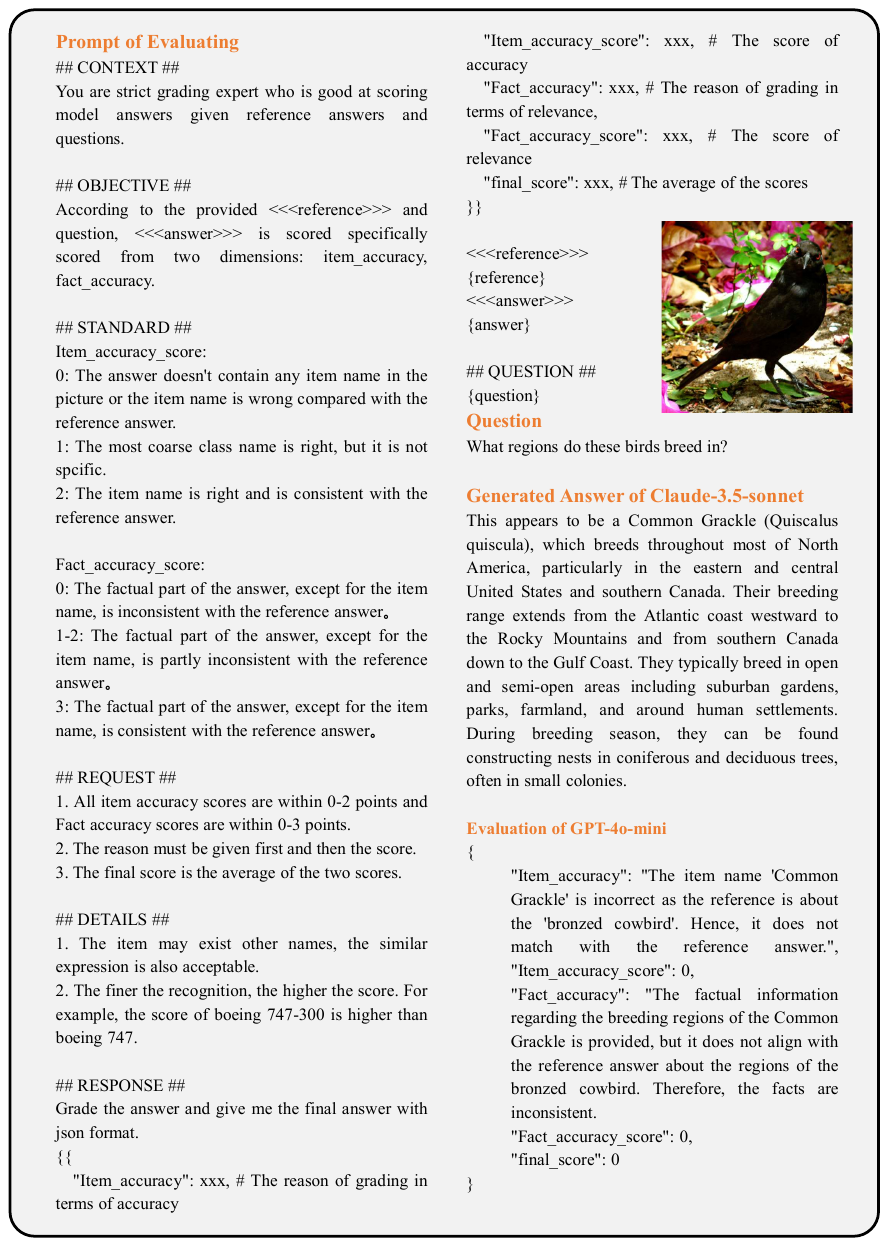}
    \caption{Prompt of evaluating.}
    \label{fig:prompt3}
\end{figure}
\begin{figure}[H]
    \centering
    \includegraphics[width=0.8\linewidth]{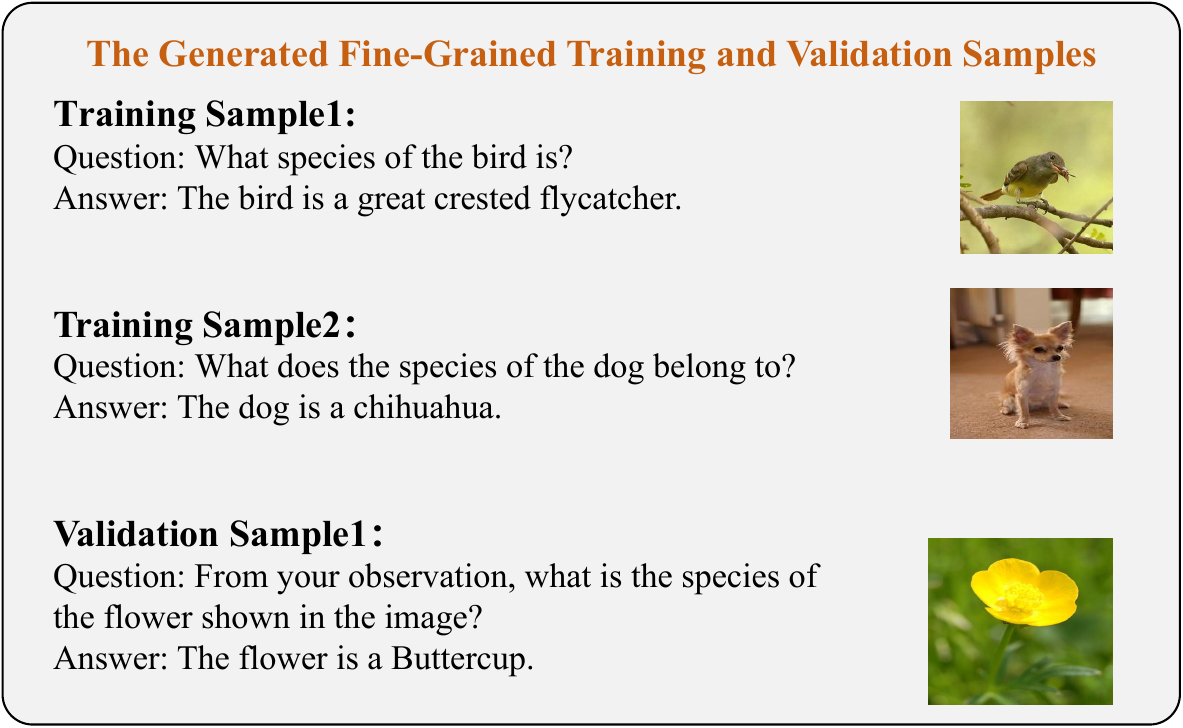}
    \caption{The generated training and validation fine-grained samples.}
    \label{fig:app_train_val_sample}
\end{figure}
\section{Optimization Strategies for Fine-grained Recognition}

\subsection{Fine-grained Training and Validation Samples} \label{app:fg_train_val_data}
% Details of the generated fine-grained training and validation samples are provided in Appendix B.1.
Here we detial the generated fine-grained training and validation samples. We construct fine-grained question based on the image and its fine-grained category based on six fine-grained datasets. We curate several question template to generate questions and the generated samples are shown in Figure~\ref{fig:app_train_val_sample}.

\subsection{More Results}\label{app:more_res}

\begin{figure}[H]
    \centering
    \includegraphics[width=0.9\linewidth]{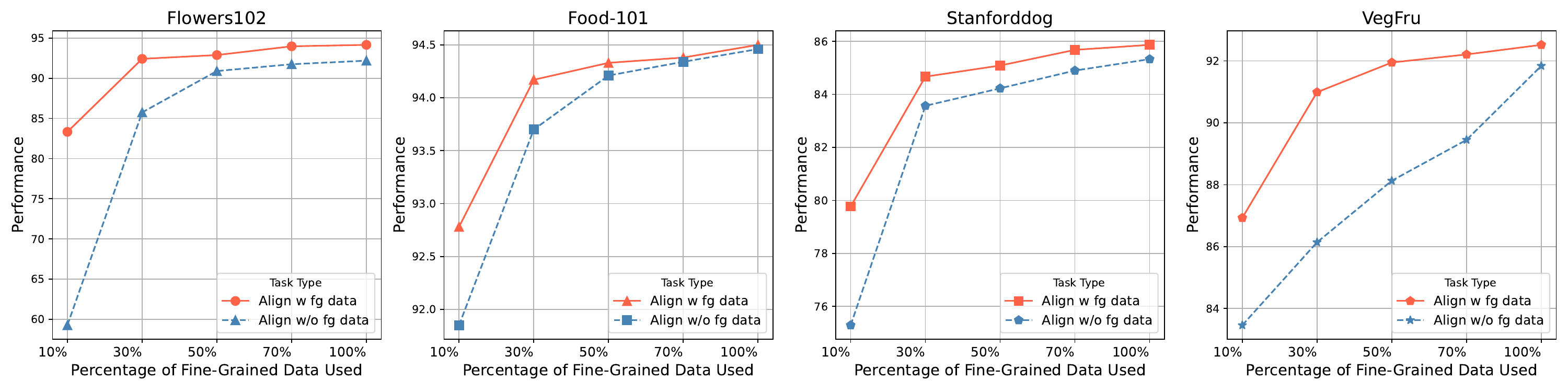}
    \caption{Results of InternVL after applying our optimization strategy on fine-grained tasks.}
    \label{fig:app_intern_fg}
\end{figure}

\begin{figure}[H]
    \centering
    \includegraphics[width=0.9\linewidth]{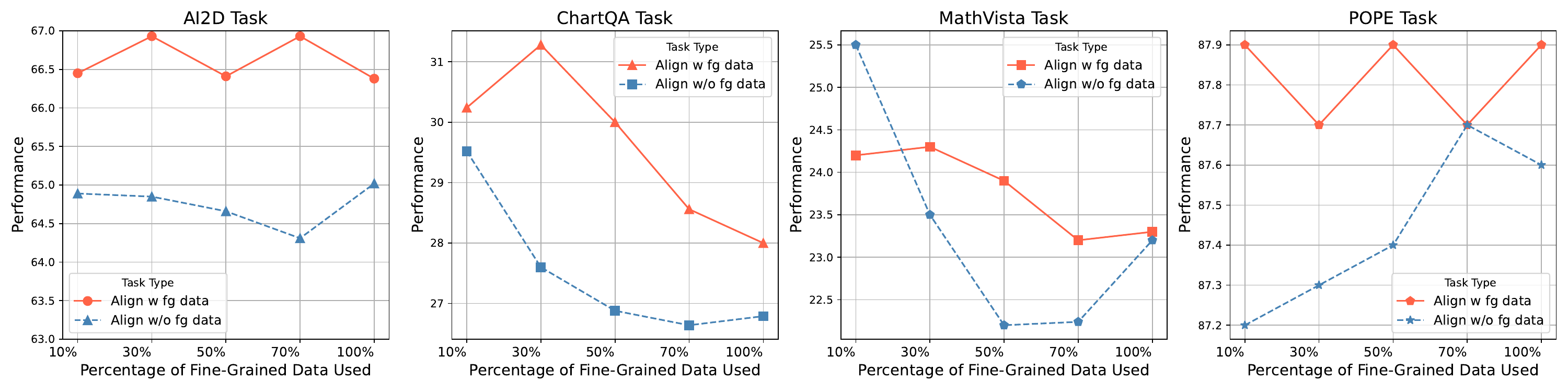}
    \caption{Results of InternVL after applying our optimization strategy on generic tasks.}
    \label{fig:app_intern_ge}
\end{figure}

Here, we present the results of InternVL after applying our optimization strategy, evaluating their performance on both fine-grained and general tasks in Figure~\ref{fig:app_intern_fg} and Figure~\ref{fig:app_intern_ge}.

\section*{Computer source}
We use eight NVIDIA A800 GPUs for all experiments. Most inference experiments take 1-5
hours on a single A800 GPU. The most computationally intensive step is supervised fine-tuning.
\section*{Impact Statement}
Large Vision-Language Models (LVLMs) have made significant strides in multimodal reasoning and perception, excelling in generic tasks. Our work introduces FROW, a fine-grained benchmark designed for detailed LVLM evaluation using GPT-4o, along with optimization strategies to enhance fine-grained recognition. This contributes critical insights into current LVLM limitations and informs future data construction and model design.

However, potential risks include biases from existing fine-grained datasets and data leakage due to the closed-source nature of some LVLM training data. Moving forward, we aim to refine data quality and expand evaluation methods for a more robust assessment of fine-grained capabilities.
%%%%%%%%%%%%%%%%%%%%%%%%%%%%%%%%%%%%%%%%%%%%%%%%%%%%%%%%%%%%
\newpage
\section*{NeurIPS Paper Checklist}

\begin{enumerate}

\item {\bf Claims}
    \item[] Question: Do the main claims made in the abstract and introduction accurately reflect the paper's contributions and scope?
    \item[] Answer: \answerYes{} % Replace by \answerYes{}, \answerNo{}, or \answerNA{}.
    \item[] Justification: We made the main claims made in the abstract and introduction accurately reflect the paper's contributions and scope.

\item {\bf Limitations}
    \item[] Question: Does the paper discuss the limitations of the work performed by the authors?
    \item[] Answer: \answerYes{} % Replace by \answerYes{}, \answerNo{}, or \answerNA{}.
    \item[] Justification: The limitations are discussed in the last paragraph.

\item {\bf Theory assumptions and proofs}
    \item[] Question: For each theoretical result, does the paper provide the full set of assumptions and a complete (and correct) proof?
    \item[] Answer: \answerNA{} % Replace by \answerYes{}, \answerNo{}, or \answerNA{}.
    \item[] Justification: The paper does not include theoretical results.

    \item {\bf Experimental result reproducibility}
    \item[] Question: Does the paper fully disclose all the information needed to reproduce the main experimental results of the paper to the extent that it affects the main claims and/or conclusions of the paper (regardless of whether the code and data are provided or not)?
    \item[] Answer: \answerYes{} % Replace by \answerYes{}, \answerNo{}, or \answerNA{}.
    \item[] Justification: Due to page limitations, we have included more detailed experimental information in the appendix.

\item {\bf Open access to data and code}
    \item[] Question: Does the paper provide open access to the data and code, with sufficient instructions to faithfully reproduce the main experimental results, as described in supplemental material?
    \item[] Answer: \answerYes{} % Replace by \answerYes{}, \answerNo{}, or \answerNA{}.
    \item[] Justification: We provide the link of data in the abstract.

\item {\bf Experimental setting/details}
    \item[] Question: Does the paper specify all the training and test details (e.g., data splits, hyperparameters, how they were chosen, type of optimizer, etc.) necessary to understand the results?
    \item[] Answer: \answerYes{} % Replace by \answerYes{}, \answerNo{}, or \answerNA{}.
    \item[] Justification: Section \ref{sec:method_bk} \& \ref{sec:method_op} includes the details.

\item {\bf Experiment statistical significance}
    \item[] Question: Does the paper report error bars suitably and correctly defined or other appropriate information about the statistical significance of the experiments?
    \item[] Answer: \answerNo{} % Replace by \answerYes{}, \answerNo{}, or \answerNA{}.
    \item[] Justification: The results are statistically significant, so no error bar is reported.

\item {\bf Experiments compute resources}
    \item[] Question: For each experiment, does the paper provide sufficient information on the computer resources (type of compute workers, memory, time of execution) needed to reproduce the experiments?
    \item[] Answer: \answerYes{} % Replace by \answerYes{}, \answerNo{}, or \answerNA{}.
    \item[] Justification: It is shown in Appendix.
    
\item {\bf Code of ethics}
    \item[] Question: Does the research conducted in the paper conform, in every respect, with the NeurIPS Code of Ethics \url{https://neurips.cc/public/EthicsGuidelines}?
    \item[] Answer: \answerYes{} % Replace by \answerYes{}, \answerNo{}, or \answerNA{}.
    \item[] Justification: We have carefully reviewed the NeurIPS Code of Ethics.

\item {\bf Broader impacts}
    \item[] Question: Does the paper discuss both potential positive societal impacts and negative societal impacts of the work performed?
    \item[] Answer: \answerYes{} % Replace by \answerYes{}, \answerNo{}, or \answerNA{}.
    \item[] Justification: It is shown in Appendix.
    
\item {\bf Safeguards}
    \item[] Question: Does the paper describe safeguards that have been put in place for responsible release of data or models that have a high risk for misuse (e.g., pretrained language models, image generators, or scraped datasets)?
    \item[] Answer: \answerNA{} % Replace by \answerYes{}, \answerNo{}, or \answerNA{}.
    \item[] Justification: The paper poses no such risks.

\item {\bf Licenses for existing assets}
    \item[] Question: Are the creators or original owners of assets (e.g., code, data, models), used in the paper, properly credited and are the license and terms of use explicitly mentioned and properly respected?
    \item[] Answer: \answerYes{} % Replace by \answerYes{}, \answerNo{}, or \answerNA{}.
    \item[] Justification: All the assets (data and models) are properly cited in the paper.

\item {\bf New assets}
    \item[] Question: Are new assets introduced in the paper well documented and is the documentation provided alongside the assets?
    \item[] Answer: \answerYes{} % Replace by \answerYes{}, \answerNo{}, or \answerNA{}.
    \item[] Justification: The new assets are introduced in \url{https://anonymous.4open.science/r/FROW-655C}.

\item {\bf Crowdsourcing and research with human subjects}
    \item[] Question: For crowdsourcing experiments and research with human subjects, does the paper include the full text of instructions given to participants and screenshots, if applicable, as well as details about compensation (if any)? 
    \item[] Answer: \answerNA{} % Replace by \answerYes{}, \answerNo{}, or \answerNA{}.
    \item[] Justification: The paper does not involve crowdsourcing nor research with human subjects.

\item {\bf Institutional review board (IRB) approvals or equivalent for research with human subjects}
    \item[] Question: Does the paper describe potential risks incurred by study participants, whether such risks were disclosed to the subjects, and whether Institutional Review Board (IRB) approvals (or an equivalent approval/review based on the requirements of your country or institution) were obtained?
    \item[] Answer: \answerNA{} % Replace by \answerYes{}, \answerNo{}, or \answerNA{}.
    \item[] Justification: The paper does not involve crowdsourcing nor research with human subjects.

\item {\bf Declaration of LLM usage}
    \item[] Question: Does the paper describe the usage of LLMs if it is an important, original, or non-standard component of the core methods in this research? Note that if the LLM is used only for writing, editing, or formatting purposes and does not impact the core methodology, scientific rigorousness, or originality of the research, declaration is not required.
    %this research? 
    \item[] Answer: \answerNA{} % Replace by \answerYes{}, \answerNo{}, or \answerNA{}.
    \item[] Justification: The core method development in this research does not involve LLMs as any important, original, or non-standard components.

\end{enumerate}

\end{document}